\documentclass[acmsmall]{acmart}

\usepackage{graphicx}
\usepackage{latexsym}
\usepackage{mathtools}

\usepackage{booktabs} 
\usepackage{xcolor}
\usepackage[linesnumbered,ruled,vlined]{algorithm2e}
\usepackage{multirow}
\usepackage{subcaption}
\usepackage{pifont}  
\usepackage{enumitem}
\setitemize{leftmargin=*,itemsep=2pt}

\usepackage{tikz}
\newsavebox{\tempbox}

\usetikzlibrary{plotmarks} 

\newcommand\marksymbol[2]{\tikz[#2,scale=1.2]\pgfuseplotmark{#1};}

\newcommand{\eg}{\textit{e.g.}}
\newcommand{\ie}{\textit{i.e.}}
\newcommand{\wrt}{\textit{w.r.t.}}
\newcommand{\method}{MANE}
\newcommand{\methodp}{\method\ensuremath{^{+}}}

\newcommand{\view}[1]{\ensuremath{^{(#1)}}}
\newcommand{\stitle}[1]{\vspace{1mm}\noindent\textbf{#1.}}

\newtheorem{definition}{Definition}

\let\vec\mathbf

\AtBeginDocument{%
  \providecommand\BibTeX{{%
    \normalfont B\kern-0.5em{\scshape i\kern-0.25em b}\kern-0.8em\TeX}}}

\setcopyright{acmcopyright}
\acmJournal{TKDD}
\acmYear{2020} 
\acmVolume{1} 
\acmNumber{1} 
\acmArticle{1} 
\acmMonth{1} 
\acmPrice{15.00}\acmDOI{10.1145/3441450}

\begin{document}

\title{Multi-View Collaborative Network Embedding}


\author{Sezin Kircali Ata}
\authornote{The author is currently affiliated with KK Women's and Children's Hospital, Singapore.}
\affiliation{%
  \institution{Nanyang Technological University}
  \streetaddress{50 Nanyang Ave}
  \country{Singapore}
  \postcode{639798}}
\email{sezin001@e.ntu.edu.sg}

\author{Yuan Fang}
\authornote{Both authors are the corresponding authors.}
\affiliation{%
  \institution{Singapore Management University}
  \streetaddress{81 Victoria St}
  \country{Singapore}
  \postcode{188065}}
\email{yfang@smu.edu.sg}

\author{Min Wu}
\authornotemark[2]
\affiliation{%
  \institution{Institute for Infocomm Research}
  \streetaddress{1 Fusionopolis Way}
  \country{Singapore}
  \postcode{138632}}
\email{wumin@i2r.a-star.edu.sg}


\author{Jiaqi Shi}
\authornote{The author is currently affiliated with University of California Irvine, United States.}
\affiliation{%
  \institution{Singapore Management University}}
\email{jqshi@smu.edu.sg}

\author{Chee Keong Kwoh}
\affiliation{\institution{Nanyang Technological University}}
\email{asckkwoh@ntu.edu.sg}

\author{Xiaoli Li}
\affiliation{\institution{Institute for Infocomm Research and Nanyang Technological University}}
\email{xlli@i2r.a-star.edu.sg}

\renewcommand{\shortauthors}{Ata and Fang, et al.}

\begin{abstract}
Real-world networks often exist with multiple views, where each view describes one type of interaction among a common set of nodes. For example, on a video-sharing network, while two user nodes are linked if they have common favorite videos in one view,  they can also be linked in another view if they share common subscribers. Unlike traditional single-view networks, multiple views maintain different semantics to complement each other. In this paper, we propose \method, a multi-view network embedding approach to learn low-dimensional representations. Similar to existing studies, \method\ hinges on diversity and collaboration---while diversity enables views to maintain their individual semantics, collaboration enables views to work together. However, we also discover a novel form of \emph{second-order} collaboration that has not been explored previously, and further unify it into our framework to attain superior node representations. Furthermore, as each view often has varying importance \wrt~different nodes, we propose \methodp, an \emph{attention}-based extension of \method\ to model node-wise view importance. Finally, we conduct comprehensive experiments on three public, real-world multi-view networks, and the results demonstrate that our models consistently outperform state-of-the-art approaches.
\end{abstract}

\ccsdesc[500]{Information systems~Data mining}
\ccsdesc[500]{Computing methodologies~Learning latent representations}
\ccsdesc[300]{Information systems~Social networks}

\keywords{multi-view networks, network embedding}


\maketitle

\section{Introduction}
\label{sec:intro}

Large-scale network data are ubiquitous in many domains, including social networks, biological networks and transportation networks.
Practical applications on these networks, such as personalized recommendations \cite{Zhang:2010a} and  
disease protein predictions \cite{ata2017disease}, 
can often be cast as link prediction and node classification tasks. Naturally, an important step towards solving these problems is to derive effective representations from the networks. 
In particular, network embedding has emerged as a promising direction and achieved considerable success, which aims to learn a low-dimensional and continuous
vector representation for each node on the network.

\begin{figure}[t]
    \centering
    \includegraphics[scale=0.5]{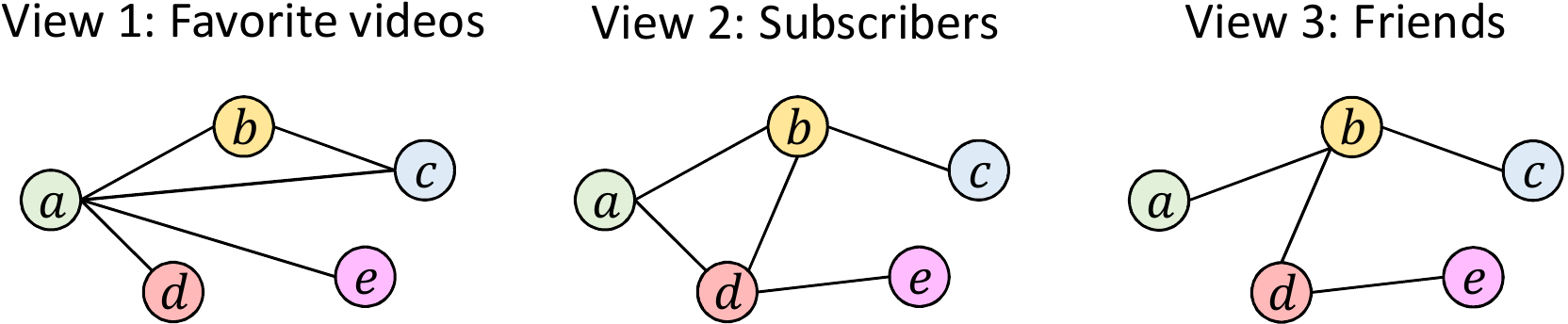}
    \caption{A toy video-sharing network with three views.}
    \label{fig:toynetwork}
\end{figure}

While many prior studies \cite{perozzi2014deepwalk,Tang2015Line} deal with single-view networks, multiple views naturally arise in real-world scenarios.
They are known as \emph{multi-view} or \emph{multiplex} networks \cite{qu2017attention,zhang2018MNE},
as shown in Fig.~\ref{fig:toynetwork}. The toy example comprises three views: the same set of users can be linked in three different ways, through sharing common favorite videos, subscribers or friends on a video-sharing platform.
Since a single view is sometimes sparse and noisy, it becomes beneficial to exploit multiple views jointly for learning more effective and robust representations. 

\begin{figure}[t]
    \centering
    \includegraphics[scale=0.5]{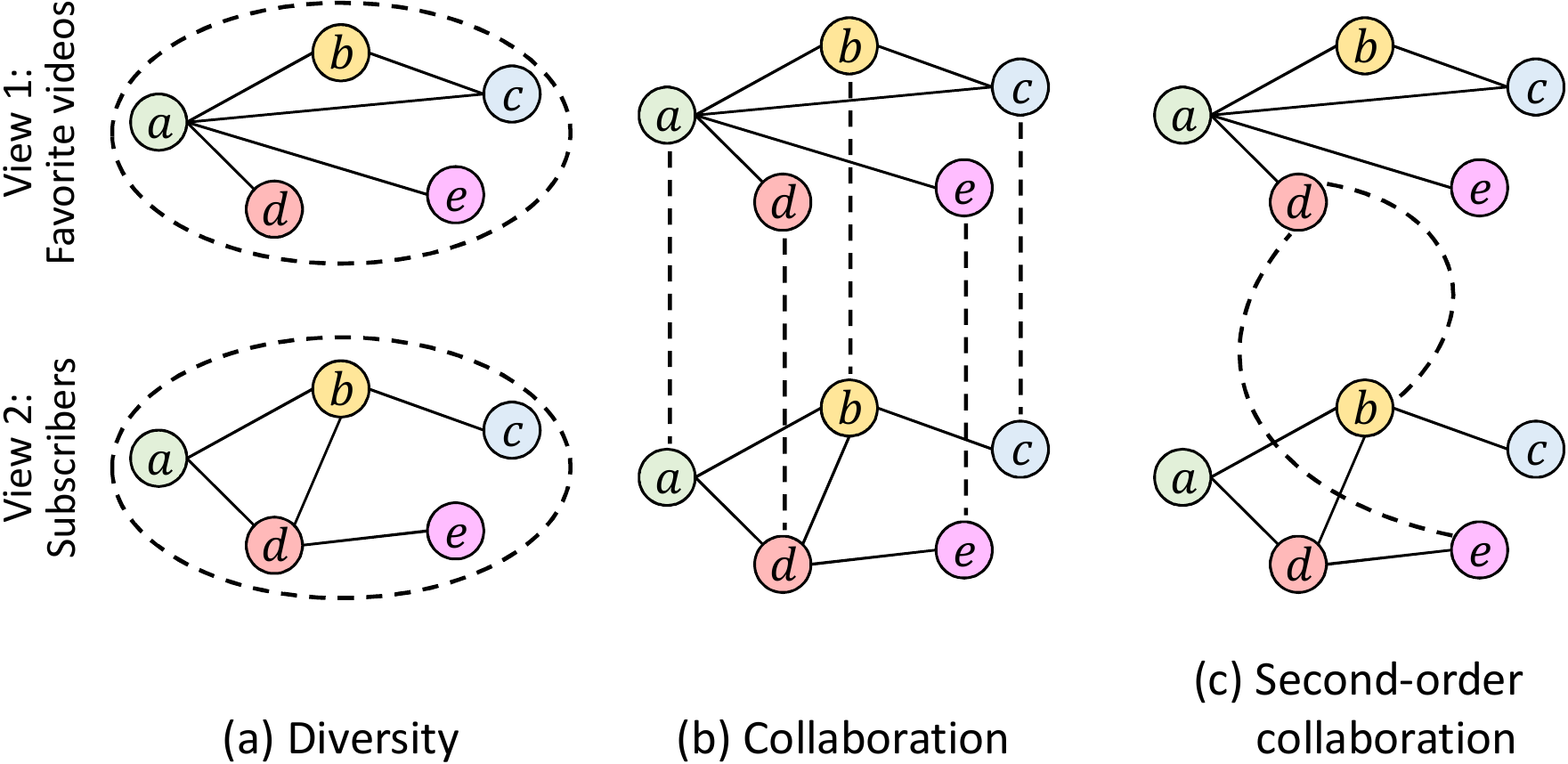}
    \caption{Three characteristics of a multi-view network.}
    \label{fig:characteristics}
\end{figure}

\stitle{Prior work} In this paper, we study the problem of multi-view network embedding. 
Although numerous multi-view learning algorithms  exist \cite{xu2013survey} 
in various contexts such as classification \cite{cao2014tensor,li2010two} and clustering \cite{chaudhuri2009multi},
fewer research have 
investigated multi-view network embedding \cite{qu2017attention,zhang2018MNE}. Existing multi-view network embedding approaches typically leverage two common \emph{characteristics} on multi-view data, namely, \emph{diversity} and \emph{collaboration}. In terms of diversity, as each view is constructed from different semantics, view-specific  representations should be learned for each view in order to retain the diverse semantics of various views. For instance, in Fig.~\ref{fig:characteristics}(a), users $a$ and $c$ are connected in view 1 as they share common favorite videos, but they have no common subscribers in view 2. Thus, $a$ and $c$'s representations should be close to one another in view 1 but not necessarily close in view 2. This can be addressed by learning one set of node representations in each view. 
In terms of collaboration, as the same node in different views ultimately describes the same instance (\eg, a particular user on a social network), its behaviors across views are not completely independent. Thus, for the same node, its  view-specific representations for different views should be synergistic and collaborate with each other, as illustrated by the cross-view alignment in Fig.~\ref{fig:characteristics}(b). 

\stitle{Our insight} While the above form of collaboration has been adopted in several state-of-the-art approaches \cite{qu2017attention,zhang2018MNE,shi2018mvn2vec}, 
it does not adequately address the collaboration of multiple views. 
Specifically, for every node, its representations across views are aligned (\ie, made close) indiscriminately, regardless of its associations (or the lack thereof) with other nodes. However, such associations are often important to multi-view collaboration---two nodes associated in one view have a better chance to also associate with each other in a different view. Our analysis of three real-world multi-view networks reveals that, two nodes linked in one view are 2.5--27.8 times more likely to also form links in a second view than two nodes not linked in the first view.\footnote{The three multi-view networks will be elaborated in our experiments in Sect.~\ref{sec:expt}, and more details of the analysis will be presented in Sect.~\ref{sec:method}.} 
In other words, two nodes associated in at least one view are more likely to lead to multi-view collaboration.
Taking Fig.~\ref{fig:toynetwork} as an example, users $d$ and $e$ have common subscribers in view 2, implying that they are likely to also ``favorite'' common videos in future, even though they are currently not linked in view 1. Thus, it is reasonable to 
pay more attention to $d$ and $e$ for multi-view collaboration. On the other hand, $b$ and $e$ are not linked in any view, which are therefore less crucial to collaboration.
We call this phenomenon the \emph{second-order} collaboration, since it deals with a pair of nodes across views. 
In contrast, the collaboration shown in Fig.~\ref{fig:characteristics}(b) is first-order, as it only deals with individual nodes.


\stitle{Present work} We aim to capture a novel form of second-order collaboration, the third characteristic of multi-view networks in addition to the previously studied diversity and first-order collaboration. 
To achieve the first-order collaboration for a given node, say $d$,  its view-specific representations are aligned across all views, as illustrated in Fig.~\ref{fig:characteristics}(b). To model the second-order collaboration for $d$, we further take account of $d$'s associations with other nodes. Using the earlier example, $d$'s representation in view 1 should exploit the knowledge that $d$ is linked to $b$ and $e$ in view 2. Unlike the first order which directly moves $d$'s representation in view 1 towards its representation in view 2, we move it towards the representations of $b$ and $e$ in view 2 to achieve the second-order multi-view collaboration, as illustrated in Fig.~\ref{fig:characteristics}(c). 
The advantage of this design is twofold. First, it explicitly exploits the associations between nodes for multi-view collaboration, which carry valuable information. Second, it results in node-wise differentiation, since the extent of collaboration a node can leverage in one view depends on its associations in other views. Both elements are neglected by the first-order collaboration in existing work.


In this paper, we unify the three characteristics of a multi-view network into one framework, and propose a new unsupervised algorithm for \underline{M}ulti-view coll\underline{A}borative \underline{N}etwork \underline{E}mbedding (\method). 
Furthermore, in a multi-view network, not all views are equally important. However, their importance to the nodes is often \emph{non-uniform}, varying from node to node. Inspired by the neural attention mechanism \cite{bahdanau2014neural}, we develop an approach called \methodp, which is an extension of \method\ to  consider node-wise view importance. The attention-based approach enables each node to locate and focus on its most important and relevant view.

We summarize the main contributions as follows. (i) We discover a new form of second-order collaboration on multi-view networks. 
    (ii) We propose a novel multi-view network embedding algorithm \method, integrating different characteristics in a unifying framework.
    (iii) We further develop \methodp\ to capture node-wise view importance through an attention mechanism. (iv) We conduct extensive experiments on three public datasets, and empirically demonstrate the superiority of our approaches.


\section{Related work}
\label{sec:relatedwork}

Network embedding has been extensively studied for its importance in real-word applications. Earlier studies \cite{perozzi2014deepwalk,Tang2015Line} have primarily devoted to preserving network structures through neighborhood sampling, such as random walks and first- or second-order proximity. More recent algorithms adopt different paradigms towards the same goal, such as GraphGAN \cite{wang2018graphgan} based on generative adversarial nets, and EP \cite{GarcaDurn2017LearningGR} based on message passing on the network. Meanwhile, graph neural networks \cite{kipf2016semi,velickovic2017graph} have recently emerged as competitive end-to-end approaches.

While the above algorithms have achieved promising results, they are only designed for single-view networks, as opposed to multi-view networks 
investigated in this paper. 
In multi-view networks, each view describes one form of interaction among a common set of nodes. This is analogous to traditional multi-view data, where each view forms a distinct feature set for a common set of instances \cite{xu2013survey}, which have been studied  in various contexts 
\cite{cao2014tensor,li2010two,greene2009matrix,liu2013multi,lu2017learning,chaudhuri2009multi,dhillon2011multi}.
Learning on multi-view networks have also been studied, such as spectral clustering \cite{kumar2011co,ma2017multi,DBLP:conf/aaai/Liu0CYRL18} and  classification \cite{he2011graphbased,wu2018multiple}. 

Recent approaches such as MVE \cite{qu2017attention}, MVNE \cite{sun2018multi}, mvn2vec \cite{shi2018mvn2vec} and MNE \cite{zhang2018MNE} intend to be more general, which aim to learn low-dimensional representations that preserve the structures of multiple views. The learned representations are often generic enough to support various applications such as node classification and link prediction. 
However, in addition to the characteristic of diversity, these multi-view network embedding methods only account for the first-order collaboration to align the representations of each node across views. 
In our approach, we discover and leverage the second-order collaboration, which promotes cross-view reinforcement based on how nodes associate with others.
Most notably, Ni \emph{et.~al.}~\cite{ni2018co} have proposed the concept of proximity disagreement, which requires the instances of the same node across views to have similar proximity to all other nodes. While this is a weaker assumption than the first-order collaboration, it is still different from our second-order collaboration as it does not leverage valuable node associations for fine-grained collaboration that varies from node to node. Moreover, a few attention-based algorithms exist \cite{qu2017attention,wang2018intra} to differentiate the importance of views in multi-view networks, 
and a more general definition of multi-view network has also been explored to model many-to-many node mappings across views \cite{ni2018co}. 
In a related but different formulation, network structures have also been studied in conjunction with content features, where the network structure can be treated as one view, and node contents as another view \cite{Huang2018unsupervised,yang2017properties,zhang2019multi}. 


On another line, network embedding algorithms have also been designed for heterogeneous information networks (HIN) \cite{sun2013mining}. A HIN consists of multiple types of nodes and edges, carrying different semantics similar to multiple views. However, HIN embedding  \cite{chang2015heterogeneous,dong2017metapath2vec,fu2017hin2vec,DBLP:conf/icdm/Zheng0LYFZTC18,hu2019hegan,zhang2020mg2vec} typically fuses and merges different types into only one view, and only  one set of node embeddings are learnt without view-specific representations. Thus, they are fundamentally single-view models,
and do not work well empirically on multi-view networks, as demonstrated in our experiments.
A recent study GATNE \cite{cen2019representation} employs a more general form of MNE \cite{zhang2018MNE} to further explore representation learning on multi-view HINs. Again, GATNE does not consider the second-order collaboration.

\section{Preliminaries}
\label{sec:prelim}



We begin with a formal definition of multi-view network, followed by the problem formulation. The main notations are listed in Table~\ref{tbl:notations}.
\begin{table}
  \caption{Summary of main notations.}
    \begin{tabular}{l|l}
    \hline
      Notation    & Definition \\
    \hline
    $U$  & the set of nodes $U$ \\
    $V$  & the set of views $V$ \\
    $E\view{v}$  & the set of edges $E\view{v}$ in view $v$ \\
    $\vec{f}_i\view{v},\tilde{\vec{f}}_i\view{v}$ & center and context embeddings of node $i$ for view $v$ \\
    $\vec{f}_i$  &  final (output) embedding of node $i$ \\
    $\Omega\view{v}$ & the set of intra-view pairs in view $v$ \\
    $i\view{v}$ & the instance of node $i$ in view $v$ \\
    $a_i\view{v}$ & attention-based weight of view $v$ \wrt~node $i$ \\
    $\vec{f}^a_i$  &  attention-based embedding of node $i$ \\
    $\alpha,\beta,\gamma$ & hyperparameters to balance loss components \\
    \hline
    \end{tabular}%
  \label{tbl:notations}%
\end{table}%

\begin{definition}
A multi-view network consists of a set of nodes $U$ and a set of views $V$, where each view $v\in V$ contains a set of edges $E\view{v}$ between the nodes in $U$.
Specifically, an edge $e\view{v}_{ij} \in E^{(v)}$ links nodes $i, j\in U$ in view $v\in V$.
We denote this multi-view network as $G=\left(U,V,\{E\view{v}:{v \in V}\}\right)$. 
\end{definition}

Given such a network $G$, the goal of multi-view network embedding is to learn a low-dimensional vector representation $\vec{f}_{i} \in \mathbb{R}^{D}$ for each node $i\in U$,
where $D$ is the dimension of the embedding space. 
More specifically, to maintain the diversity of views, for each node $i\in U$ and each view $v\in V$, we learn an intermediate view-specific representation $\vec{f}_{i}\view{v} \in \mathbb{R}^{\left\lfloor{D/|V|}\right\rfloor}$.
The final representation $\vec{f}_{i}$ is aggregated from these intermediate representations: 
\begin{align}
\vec{f}_i=\oplus_{v \in V} \vec{f}_i\view{v},
\end{align}
where $\oplus$ denotes some form of aggregation, such as vector concatenation as adopted in our formulation. 
The final representations can be used as features in a variety of applications such as node classification and visualization, relationship mining and link prediction. 

\section{Proposed Approach}
\label{sec:method}

In this section, we introduce our models \method\ and \methodp\ for multi-view network embedding.

\subsection{Overall Framework}

\begin{figure}[t]
\includegraphics[scale=0.5]{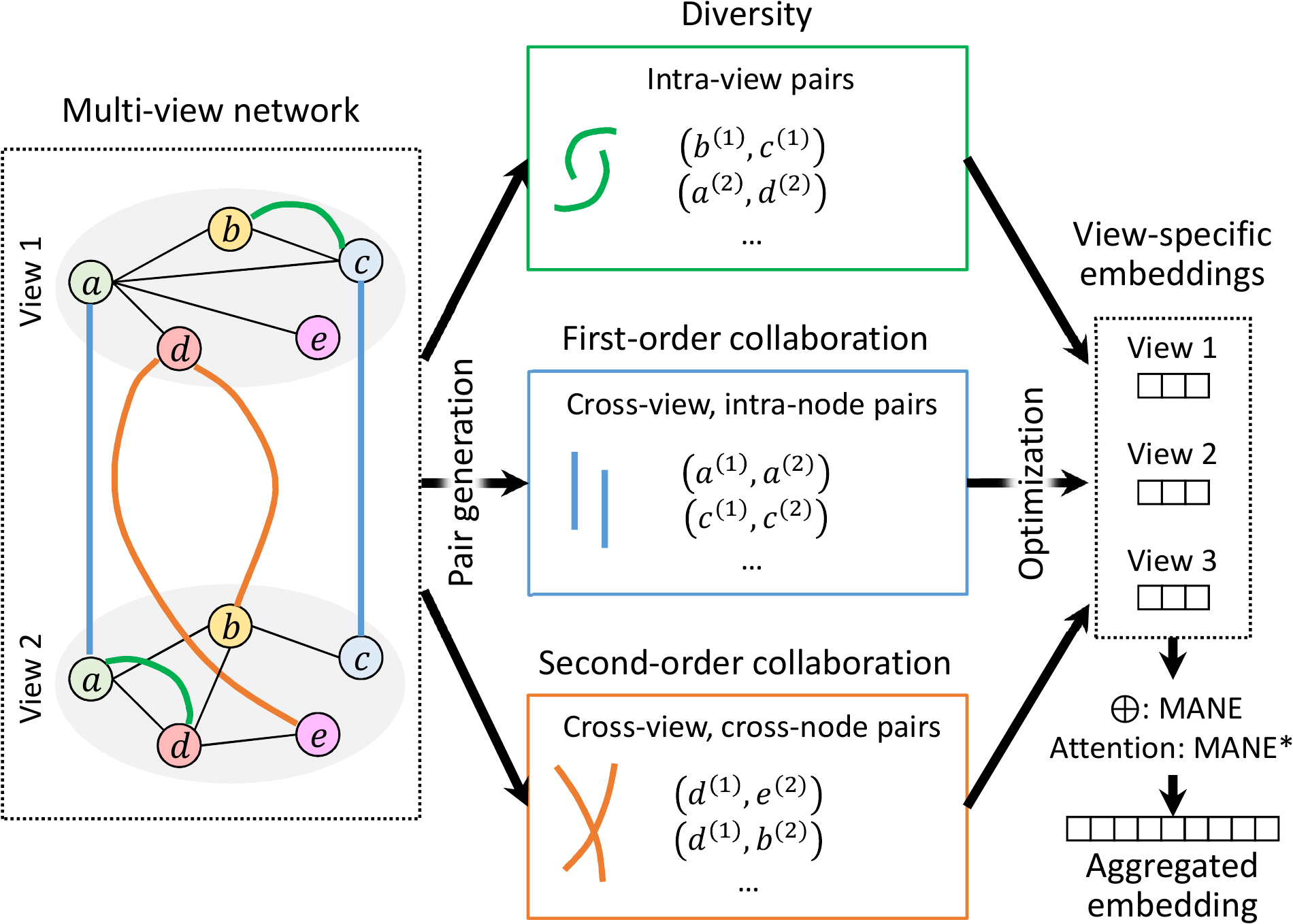}
\caption{Overall framework of \method.}
\label{fig:overall-framework}
\end{figure}

To start, we present the overall framework in Fig.~\ref{fig:overall-framework}. 
The three characteristics of a multi-view network can be unified by three categories of node pairs.

In the first category, we consider \emph{intra-view} pairs, which can be generated using random walks in each view \cite{perozzi2014deepwalk}. 
By modeling intra-view pairs in each view, we are able to capture the diversity of different views.
In the second category, we consider \emph{cross-view, intra-node} pairs,
where instances of the same node (\ie, intra-node) across two views (\ie, cross-view) form pairs. By aligning node representations in such a pair, the first-order collaboration can be captured. 
In the third category, we consider \emph{cross-view, cross-node} pairs,
where a node in one view form pairs with different nodes (\ie, cross-node) in another view (\ie, cross-view) based on their associations in each view, to capture the second-order collaboration.

The three categories of node pairs are jointly trained to derive one set of embeddings for each view, which are further aggregated to produce the final embeddings. In particular, during aggregation, \method\ assumes equal weight across views, and \methodp\ learns a different weight for each view.  


\subsection{Characteristics of Multi-View Networks}

In the following, 
we delve into each characteristic and its corresponding node pairs.

\stitle{Diversity: Intra-view pairs} 
To retain the diversity of each view, 
each node has a view-specific representation such that two nodes associated in the same view should have similar representations for that view. Following the DeepWalk model \cite{perozzi2014deepwalk}, for each view, we generate topologically associated node pairs from random walks. Such pairs are also intra-view, as each pair is generated within one view. As shown in Fig.~\ref{fig:overall-framework}, we consider intra-view pairs such as $(b\view{1},c\view{1})$ in view 1 and $(a\view{2},d\view{2})$ in view 2, 
where the superscripts indicate the views from which the node instances are observed. 

Formally, for a given view $v\in V$, we assume a set of intra-view pairs $\Omega\view{v} \subset U \times U$. 
Adopting the well-known skipgram model \cite{perozzi2014deepwalk}, a pair $(i\view{v},j\view{v}) \in \Omega\view{v}$ consists of a center node $i\view{v}$ and a context node $j\view{v}$. Given a center node, the task is to predict its context node, \ie, to maximize $P(j\view{v}|i\view{v})$. 
Subsequently, we optimize the following loss for some model parameters $\Theta$: 
\begin{align}
L_\text{Div}(\Theta)=  -\sum_{v\in V} \sum_{(i\view{v},j\view{v})\in \Omega\view{v}}\log P(j\view{v}|i\view{v};\Theta).
\label{eq:intra-view-cross-node}
\end{align}
Here $P(j\view{v}|i\view{v};\Theta)$ is further defined with a softmax function:
\begin{align}
P(j\view{v}|i\view{v};\Theta) = \frac{\exp\left(\tilde{\vec{f}}_j\view{v} \cdot \vec{f}_{i}\view{v}\right)}{\sum_{u \in U} \exp\left(\tilde{\vec{f}}_{u}\view{v} \cdot \vec{f}_{i}\view{v}\right)},
\label{eq:conditional_prob}
\end{align}
where $\vec{f}_i\view{v},\tilde{\vec{f}}_i\view{v}\in \mathbb{R}^{\left\lfloor{D/|V|}\right\rfloor}$ respectively denote the center and context embeddings of node $i$ specific to view $v$.
Thus, the model parameters consist of the center and context embeddings of all nodes in all views, \ie, 
$\Theta=\{\vec{f}_i\view{v},\tilde{\vec{f}}_i\view{v}:(i,v)\in U\times V\}$.
Essentially, the views are decoupled from each other to capture the diversity of each view, as no cross-view pairs are involved.

\stitle{First-order collaboration: Cross-view, intra-node pairs}
While different views exhibit diversity, they ultimately converge on a common set of nodes. Since the instances of the same node across different views fundamentally describes the same entity, such as a particular user on a social network, its view-specific representations should  collaborate with one another. As shown in Fig.~\ref{fig:overall-framework}, we consider  cross-view, intra-node pairs such as $(a\view{1},a\view{2})$
and $(c\view{1},c\view{2})$, which can be generated across different views for the same node. In each pair, the two view-specific embeddings of the same node should become similar, by optimizing the following loss: 
\begin{align}
L_\text{C1}(\Theta) &=  -\sum_{v\in V} \sum_{(i\view{v},\cdot)\in \Omega\view{v}} \sum_{v'\ne v} \log P(i\view{v'}|i\view{v};\Theta)\nonumber\\
&=  -\sum_{v\in V} \sum_{(i\view{v},\cdot)\in \Omega\view{v}} \sum_{v'\ne v}\log  \frac{\exp\left(\vec{f}_i\view{v'} \cdot \vec{f}_{i}\view{v}\right)}{\sum_{u \in U} \exp\left(\vec{f}_{u}\view{v'} \cdot \vec{f}_{i}\view{v}\right)}.\label{eq:cross-view-intra-node}
\end{align}
 
Note that we form a cross-view, intra-node pair $(i\view{v},i\view{v'})$ for the center node $i$ in every intra-view pair $(i\view{v},\cdot)\in \Omega\view{v}$.
Thus, a frequently observed center node in a view, which has more topological importance in that view, will also be updated more in the first-order collaboration
to ensure a balanced optimization.

\stitle{Second-order collaboration: Cross-view, cross-node pairs}
As motivated in Sect.~\ref{sec:intro}, our key intuition is that two nodes linked in one view are often more likely to also link with each other in another view. 
Given two views $v \neq v'$, this intuition can be quantified by the following ratio:
\begin{align}
\frac{P\left(e_{ij}\view{v'} \in  E\view{v'} | e_{ij}\view{v} \in E\view{v}\right) }{P\left(e_{ij}\view{v'} \in  E\view{v'}| e_{ij}\view{v} \notin E\view{v}\right)}.
\end{align}
The larger the ratio, the more likely that two nodes linked in one view will also form a link with each other in another view;
a ratio of 1 would imply that whether two nodes are linked in one view has no bearing on their link formation in another view.
On the three real-world multi-view networks used in our experiments, namely Alzheimer's, LinkedIn and YouTube (details in Sect.~\ref{sec:expt}), the above ratio for all pairwise views ranges from 2.5 to 27.8 with a median of 7.0 as shown in Table~\ref{tab:pairwiseRatio}.
That means on these networks, node pairs linked in one view are significantly more likely than those not linked to also form links in another view. Thus, the second-order collaboration is a fair assumption on many multi-view networks, although it is not universal. As we shall see later, our model is flexible to adjust the contribution from the second-order collaboration.

\begin{table}[tbp]

  \caption{Second-order collaboration between pairwise views ($v,v'$) in three multi-view networks.}

\begin{subtable}{.5\linewidth}
{  \[\begin{array}{r|c|c}
 \hline
v\setminus{v'}
& \multicolumn{1}{c|}{\text{GO}} & \multicolumn{1}{c}{\text{PPI}}   \\
\cline{1-3}    \text{GO}    & -  & \multicolumn{1}{c}{27.8}   \\
\cline{1-3}    \text{PPI}   & 27.7  & \multicolumn{1}{c}{-}   \\
    \hline
\end{array}\]
}
\caption{Alzheimer's}\label{tab:pairwiseRatio1}
\end{subtable}
\begin{subtable}{.5\linewidth}
{\[\begin{array}{r|c|c|c}
    \hline
     v \setminus{v'}
    & \multicolumn{1}{l|}{\text{College}} & \multicolumn{1}{l|}{\text{Employer}} & \multicolumn{1}{l}{\text{Friend}} \\
        \hline
        \text{College} & -  & 6.4   & 27.8 \\
        \hline
        \text{Employer} & 6.6   & -   & 12.6 \\
        \hline
        \text{Friend} & 15.7  & 9.1   & -\\
            \hline
        \end{array}\]
\caption{LinkedIn}\label{tab:pairwiseRatio2}
}
\end{subtable}    
 \begin{subtable}{.5\linewidth}
{\[\begin{array}{r|c|c|c}   
    \hline

    v\setminus{v'}
    & \multicolumn{1}{l|}{\text{Friend}} & \multicolumn{1}{l|}{\text{Subscriber}} & \multicolumn{1}{l}{\text{Video}} \\
        \hline
        \text{Friend} & -   & 6.6   & 4.5 \\
        \hline
        \text{Subscriber} & 7.5   & -   & 2.5 \\
        \hline
        \text{Video}  & 5.1   & 2.6   & - \\
        \hline
    \end{array}\]
}
\caption{YouTube}\label{tab:pairwiseRatio3}
\end{subtable}   
\label{tab:pairwiseRatio}
\end{table}%

Using Fig.~\ref{fig:overall-framework} as an example, nodes $d$ and $e$ are topologically related in view 2 but not in view 1. Under the second-order collaboration, the two nodes may entail a latent, unknown association in view 1, which is not currently manifested, possibly due to incomplete and noisy data, or a time delay---they may develop an explicit link in future. It is important to note that this assumption merely suggests a tendency. Thus, it is not appropriate to directly or explicitly make $d$ and $e$'s representations closer in view 1. 
Instead, we propose an implicit mechanism to enable such collaboration across views. As shown in Fig.~\ref{fig:overall-framework}, we consider cross-view, cross-node pairs like $(d\view{1}, e\view{2})$ and  $(d\view{1}, b\view{2})$. That is, we move $d$'s representation in view 1
towards $e$ and $b$'s in view 2, enabling the two views to collaborate depending on $d$'s associations to other nodes.
In contrast, the first-order collaboration simply make $d$'s representations in both views closer without the knowledge of $d$'s associations.

Formally, 
suppose there exists an intra-view pair in view $v$, say $(i\view{v},j\view{v})\in \Omega\view{v}$, which means node $i$ is topologically associated with $j$ in view $v$.
Given such an association, we can form a cross-view, cross-node pair $(i\view{v'},j\view{v})$, to move $i$'s representations in another view $v'$ towards $j$'s representation in view $v$. Thus, we optimize the following loss:
\begin{align}
L_\text{C2}(\Theta)&= - \sum_{v\in V} \sum_{(i\view{v},j\view{v})\in \Omega\view{v}}\sum_{v'\ne v}\log P(i\view{v'},j\view{v};\Theta) \nonumber\\
&= - \sum_{v\in V} \sum_{(i\view{v},j\view{v})\in \Omega\view{v}} \sum_{v'\ne v} \frac{\exp\left(\tilde{\vec{f}}_j\view{v} \cdot \vec{f}_{i}\view{v'}\right)}{\sum_{u \in U} \exp\left(\tilde{\vec{f}}_{u}\view{v} \cdot \vec{f}_{i}\view{v'}\right)}. \label{eq:cross-view-cross-node}
\end{align}

\subsection{The \method\ Algorithm}
\label{sec:model:algorithm}

With the losses based on the three categories of node pairs, we further present the proposed approach \method.




\stitle{Loss function}
We combine the losses of the three characteristics in Eq.~\eqref{eq:intra-view-cross-node}, \eqref{eq:cross-view-intra-node} and \eqref{eq:cross-view-cross-node} into the following, as the overall loss for a multi-view network. 
\begin{align}
    L =  L_\text{Div} + \alpha \cdot L_\text{C1} + \beta \cdot L_\text{C2},\label{eq:overall_loss}
\end{align}
where $\alpha,\beta \ge 0$ are hyperparameters to control the relative importance among the three components.
This is an unsupervised formulation, where we  maximize the likelihood of the parameters $\Theta$
only based on the three categories of node pairs, and produce final embeddings as features for downstream applications. However, depending on the application, supervision may still be leveraged, such as node class labels in node classification.

As a further remark, when $\alpha=\beta=0$, \method\ reduces to Skip-gram models such as DeepWalk \cite{perozzi2014deepwalk} on individual views in a decoupled manner, without leveraging any collaboration between views. When $\alpha>0$ and $\beta=0$, \method\ only captures diversity and first-order collaboration.

\stitle{Algorithm}
Our approach is summarized in Algorithm \ref{Algorithm1}. We first initialize the parameters randomly (line 1).
Next, for each view $v$, we sample random walks starting from every node, and further generate intra-view node pairs $\Omega\view{v}$ based on the sampled random walks (lines 2--4). 
In the succeeding block, we perform gradient updates on the parameters (lines 5--14). More specifically, we form intra-view node pairs to capture diversity (lines 8--9),
cross-view, intra-node pairs to capture first-order collaboration (lines 11--12), and cross-view, cross-node pairs to capture second-order collaboration (lines 13--14).  
Finally, we concatenate the view-specific center embeddings of each node to obtain the final embedding (lines 15--16).

To implement the algorithm, we follow DeepWalk \cite{perozzi2014deepwalk} to perform truncated random walks in each view, as well as to further generate center-context node pairs from the sampled random walks. Such center-context pairs form the intra-view pairs. We also adopt negative sampling for the softmax function in each loss component to accelerate the training process. To minimize the overall loss function, we employ the Adam optimizer.

\begin{algorithm}[t]
    \SetArgSty{textnormal}
    \SetKwInOut{Input}{Input}
    \SetKwInOut{Output}{Output}

    \Input{$G=\left(U,V,\{E\view{v}:{v \in V}\}\right)$, $\alpha$, $\beta$ }
    \Output{final embeddings $\left\{\vec{f}_i:i\in U\right\}$}
    Randomly initialize parameters $\Theta$\\
    \For {\text{each view} $v \in V$}{
     $R\view{v} = \text{SampleRandomWalks}\,(U, E\view{v})$\\
     $\Omega\view{v} = \text{GenerateIntraViewPairs}\,(R\view{v})$\\
    }
    
    \While {not converged} {
        \For {each view $v \in V$} {
            Shuffle pairs in $\Omega\view{v}$ \\
            \For {each intra-view pair $(i\view{v},j\view{v}) \in \Omega\view{v}$}{
               Update $\vec{f}_i\view{v},\tilde{\vec{f}}_j\view{v},\tilde{\vec{f}}_u\view{v}$ based on  Eq.~\eqref{eq:intra-view-cross-node}\\
               \For {each view $v' \ne v$}{
                    Form a cross-view, intra-node pair $(i\view{v},i\view{v'})$\\
                    Update $\vec{f}_i\view{v},\vec{f}_i\view{v'},\vec{f}_u\view{v'}$ based on Eq.~\eqref{eq:cross-view-intra-node}\\
                    Form a cross-view, cross-node pair $(i\view{v'},j\view{v})\hspace{-2mm}$\\
                    Update $\vec{f}_i\view{v'},\tilde{\vec{f}}_j\view{v},\tilde{\vec{f}}_u\view{v}$ based on  Eq.~\eqref{eq:cross-view-cross-node}\\
               }
            }
          }
        }
    \For {$i \in U$}{
        $\vec{f}_i=\oplus_{v \in V} \vec{f}_i\view{v}$\\
    }
    \Return{$\left\{\vec{f}_i:i\in U\right\}$}\\
    \caption{Training of \method}
\label{Algorithm1}
\end{algorithm}

\stitle{Complexity analysis}
Based on existing work \cite{qu2017attention}, learning view-specific embeddings over all views takes $O\left(|E|\cdot D/|V|\cdot K\right)$ time, where $|E|=\sum_{v \in V} |E\view{v}|$ denotes the total number of edges in all views, $\left\lfloor{D/|V|}\right\rfloor$ is the dimension for view-specific embeddings, and $K$ is the number of negative samples. This is equivalent to the time complexity of intra-view-cross-node consistency. 
On the other hand, the complexities for cross-view-intra-node and cross-view-cross-node consistencies are both  $O\left(|E|\cdot D/|V|\cdot K\cdot|V|\right)$. Thus, the overall time complexity is $O(|E|\cdot D\cdot K)$. Thus, \method\ scales linearly in the total number of edges in all views. 

\subsection{Extension with View Attention}

In \method, we treat all views with equal importance during the final aggregation. However, in real-world scenarios, the importance or relevance of each view is often non-uniform, and varies from node to node. Thus, we propose \methodp, an extension of \method\ to account for node-wise view importance. 

\stitle{View attention}
To allow a node to focus on the view most important to it, we adopt the neural attention mechanism \cite{bahdanau2014neural} 
to model view attention. We first define 
a score function  $s:U\times V \to \mathbb{R}$ to compute the similarity between any node $i$ and any view $v$:
\begin{align}
s(i,v) =\vec{z}_2\view{v} \cdot \tanh (\vec{z}_1\view{v} \circ \vec{f}_{i}+b_1\view{v}\vec{1})+b_2\view{v},
\end{align}
where $\vec{z}_1\view{v},\vec{z}_2\view{v}\in \mathbb{R}^D,b_1\view{v},b_2\view{v}\in \mathbb{R}$ are trainable parameters specific to view $v$,
and $\circ$ denotes the Hadamard product. 
The scores are further normalized to obtain the attention value ${a_i\view{v}}$, \ie, the weight of view $v$ \wrt\ node $i$:
\begin{align}
{a_i\view{v}}=\frac{\exp s(i,v)} {\sum_{v' \in V} \exp s(i,v')}.
\end{align}
Subsequently, for every node $i$, we  aggregate its view-specific embeddings into a vector $\vec{f}_i^a \in \mathbb{R}^D$,  weighted by its view attention:
\begin{align}
\vec{f}^a_i = \tanh \left( \oplus_{v \in V} a_{i}\view{v}\vec{f}_i\view{v}  \right).
\end{align}
The attention-based embeddings, $\{\vec{f}^a_i:i\in U\}$, can be used as features for downstream applications, where some supervision is often available to guide the learning of view-specific attention parameters \cite{qu2017attention}. For example, we could minimize the following loss for node classification:
\begin{align}
L_\text{Att} =  \frac{1}{|U_\text{train}|}\sum_{i\in U_\text{train}} \text{CrossEntropy}(y_i,h(\vec{f}_i^a)),
\end{align}
where $U_\text{train}$ is the set of training nodes, $y_i$ is the class of node $i$, and $h(\vec{f}_i^a)$ is the classification output for node $i$ using the attention-based embedding   $\vec{f}_i^a$. Since $U_\text{train} \subset U$, this is a semi-supervised setting.
Similarly, for link prediction an alternative classification function $h'(\vec{f}_i^a,\vec{f}_j^a)$ could be employed, using the attention-based embedding of two nodes.  

\stitle{Overall loss}
We formulate the loss of \methodp\ by extending the loss of \method\ in Eq.~\eqref{eq:overall_loss}, to include the attention component:
\begin{align}
L^+ = L_\text{Div} + \alpha \cdot L_\text{C1} + \beta \cdot L_\text{C2} +\gamma \cdot L_\text{Att},
\end{align}
where $\gamma>0$ is a hyperparameter to control the contribution of the attention-based loss.

Due to the additional loss component, compared to \method, an extra time complexity of $O(T\cdot D \cdot |V|)$ will be added to \methodp, where $T$ is the size of the training data.
Since \method\ is linear in the number of edges $|E|$, and $T$ is typically much smaller than $|E|$ for node classification and comparable in scale to $|E|$ for link prediction,
the extra complexity term does not present a significant overhead.

\section{Experiments}
\label{sec:expt}

We evaluate the performance of our proposed methods and conduct an in-depth model analysis on three real-world multi-view networks. 
Results demonstrate the superior performance of \method\ and \methodp\ against a comprehensive suite of state-of-the-art baselines. 

\subsection{Experimental Setup}

\stitle{Datasets and tasks}
Table~\ref{tab:datasets} summarizes three public, real-world multi-view networks used in our experiments, as follows.

\begin{table}
  \centering
  \caption{Summary of datasets.}
\begin{tabular}{c|c|c|c|c} 
\hline
Dataset & \# Nodes & \multicolumn{3}{c}{\# Edges in each view} \\\hline
\multirow{2}{*}{Alzheimer's} & \multirow{2}{*}{12\,901} & PPI & GO  & \\
       &       & 96\,845 & 107\,508  \\\hline
\multirow{2}{*}{LinkedIn} & \multirow{2}{*}{10\,196} & College & Employer & Friend \\
       &       & 1\,527\,681 & 1\,765\,927 & 29\,434 \\\hline
\multirow{2}{*}{YouTube} & \multirow{2}{*}{7\,558} & Friend & Subscriber & Video \\
       &       & 1\,003\,923 & 1\,547\,091 & 1\,918\,208 \\\hline
\end{tabular}%
  \label{tab:datasets}%
\end{table}%

\textbf{\emph{Alzheimer's}}: A protein network to identify causative genes for Alzheimer's disease, which includes two views, namely, PPI and GO. The PPI view captures the protein-protein interactions from the IntAct database \cite{IntAct}, whereas the GO view captures the functional associations between proteins based on their Gene Ontology (GO) annotations \cite{wang2007new}. 
For the GO view, we first calculated the pairwise functional similarity between proteins, and subsequently built a K-NN similarity graph with $K=10$. In this two-view protein network, each node has a binary label to show whether it is a causative gene for Alzheimer's disease. In other words, the task of disease gene identification for Alzheimer's can be cast as a binary classification problem. In particular, this is a highly skewed dataset with fewer than 1\% positives. Thus, we have also tried to over-sample with SMOTE \cite{chawla2002smote}, although it yielded similar results.




\textbf{\emph{LinkedIn}} \cite{linkedin}: 
A professional social network with three views, namely, college, employer and friend. Edges in the college or employer view indicate common college or workplace between users, respectively. Edges in the friend view represent LinkedIn connections. 
Our task deals with relationship mining 
to uncover latent relationship types between two nodes. Some of the node pairs have a label to indicate their relationship, such as personal community or professional contact. 
In total there are eight classes of relationship, and is thus a multi-class problem. Note that the relationship classes are latent and are independently labeled by the users, which do not directly correspond to existing edges in any of the three views. 


\textbf{\emph{YouTube}} \cite{youtube}: 
A video-sharing network with three views, namely, friend, subscriber and video, where two users are linked if they have common friends, subscribers and favorite videos, respectively. Our task is to perform link prediction on an additional contact network, 
and can be formulated as a binary classification problem. 
Note that the friend view excludes all the contacts from the ground truth contact network. To provide negative instances, we randomly sample five times as many non-contact users for each positive contact.


 \stitle{Evaluation metrics}
For binary classification on Alzheimer's and YouTube, we employ areas under the ROC and Precision-Recall (PR)  curve;
for multi-class classification on LinkedIn, we use micro- and macro-F scores. 

\stitle{Baselines} We compare with several common strategies (Single, Decoupled and Merged), as well as state-of-the-art methods for heterogeneous information network embedding (HIN2Vec and HeGAN) and multi-view network embedding (MVE, mvn2vec, MNE and GATNE).

\textbf{\emph{Single}}: A Skip-gram model equivalent to  DeepWalk applied to a single view. 
For each task, we run the model on every view, and report the average performance.

\textbf{\emph{Decoupled}}: Each view is independently trained by the single-view Skip-gram model, with the number of dimensions set to $\left\lfloor{D/|V|}\right\rfloor$. All views' embeddings are then concatenated to form the final embedding for each node. 

\textbf{\emph{Merged}}: All views are merged into one network by taking the union of the edges in all views. The single-view Skip-gram model is then trained on the merged network. 

\textbf{\emph{HIN2Vec}} \cite{fu2017hin2vec}: A heterogeneous network embedding approach, which samples various meta-paths and feeds them into a neural network. We merged all views to form a heterogeneous graph, where edges from different views belong to different types of relation.

\textbf{\emph{HeGAN}} \cite{hu2019hegan}: A heterogeneous network embedding approach, which employs the adversarial principle to learn more robust  relationships between nodes. Similar to HIN2Vec, it works on a heterogeneous graph as constructed in the above. 

\textbf{\emph{MVE}}  \cite{qu2017attention}: A multi-view network embedding approach, which promotes the first-order collaboration by regularizing the distance between view-specific embeddings and the final embedding. 
It also has a semi-supervised variant to learn view attention, which we denote as MVE$^+$. 

\textbf{\emph{mvn2vec}} \cite{shi2018mvn2vec}: A multi-view embedding approach with two variants \textbf{\emph{mvn2vec-c}} and \textbf{\emph{mvn2vec-r}}.
To enforce first-order collaboration, mvn2vec-c variant employs partial parameter sharing across views, 
whereas mvn2vec-r regularizes the distance between view-specific embeddings. 

\textbf{\emph{MNE}}  \cite{zhang2018MNE}: A multi-view network embedding algorithm. For each node, the model learns a final embedding consisting of a common embedding for first-order collaboration, and view-specific relation-based embeddings. 

\textbf{\emph{DMNE}}  \cite{ni2018co}:  A multi-view network embedding algorithm with potentially many-to-many node mappings across views, although in our setting only one-to-one mappings exist. We adopted their proximity disagreement formulation, due to its weaker assumption and better empirical performance. 

\textbf{\emph{GATNE}} \cite{cen2019representation}: An embedding algorithm for attributed multi-view heterogeneous network based on MNE, although in our setting we only consider one node type with multiple edge types (\ie, views).
Moreover, we used their transductive version since we do not consider node attributes in our setting.
We compare to its variants both with and without the attention mechanism, and denote the attention-based version as GATNE$^+$. 

\stitle{Implementation details}
To sample random walks for all Skip-gram models, we applied walk length of 10, 5 walks per node, windows size of 3, and 10 samples per negative sampling. All methods adopt $D=128$ as the dimension of the final embedding. 
For the baselines, we used the implementations from their respective authors, and
extensively tuned their main hyperparameters. Specifically, we chose $\eta=0.05$ for MVE, $\theta=0.8$ for mvn2vec-c and $\gamma=0.01$ for mvn2vec-r, and $r=1000$ for MNE, which give competitive results empirically.
Our methods are implemented in Python using PyTorch. We set $\alpha=\beta=1$ for an equal weight between the three characteristics. We further set the attention hyperparameter $\gamma=1000$ for LinkedIn and YouTube, which is generally robust and insensitive to small changes. On the highly skewed Alzheimer's dataset, we set a smaller $\gamma=0.1$ to reduce reliance on the very skewed training nodes. Nonetheless, we also conducted a sensitivity study to show the impact of these hyperparameters. 
Finally, for all methods, the learned final embeddings are used as features to further train a logistic regression model using five-fold cross-validation.
For the semi-supervised attention-based methods, only the same training folds from the downstream application is utilized during the training of node embeddings.

\begin{table}[tbp]
  \centering
    \caption{Performance evaluation on three datasets without view attention (bold: best; underline: runner-up).}  
    \label{tab:MICROS_performance}%
\begin{tabular}{l|cc|cc|cc}
     \hline
          & \multicolumn{2}{c|}{Alzheimer's} & \multicolumn{2}{c|}{LinkedIn} & \multicolumn{2}{c}{YouTube} \\
     \hline
     Metric  &  ROC-AUC    &  PR-AUC     &  Micro-F  &  Macro-F  &  ROC-AUC    &  PR-AUC     \\
     \hline
     Single  & 0.6968 & 0.0221 & 0.4001 & 0.3468 & 0.6334 & 0.1565 \\
     Decoupled  & 0.8200 & 0.0735 & 0.4341 & 0.3739 & 0.6700 &  {0.1649} \\
     Merged  & 0.7305 & 0.0075 & 0.4197 & 0.3724 & 0.6565 &  \underline{0.1763}     \\
     \hline
     HIN2Vec  & 0.5734 & 0.0030 & 0.3014 & 0.2175 & 0.6264 & 0.1258 \\
     HeGAN  & 0.6967 & 0.0104 & 0.3937 & 0.3467 & 0.6322 & 0.1520 \\
     \hline
     MVE  & 0.6543 & 0.0161 & 0.4113 & 0.2867 & 0.6276 & 0.1650 \\
      mvn2vec-c  & 0.8617 &  {0.0890}  & 0.4326 & \underline{0.3791}   & 0.6633 & 0.1621 \\
     mvn2vec-r   & 0.8756 & 0.0275 & \underline{0.4439}   & 0.3717 & {0.6703}     & 0.1669 \\
     MNE  &  {0.9195} & \underline{0.1676} &   {0.4334} & 0.3538 & {0.6749}  & {0.1604}  \\
    DMNE  &\underline{0.9357}  & 0.0603 & 0.3473 & 0.2253 & 0.6056 & 0.1394 \\
    GATNE  & 0.9190 & 0.1227 & 0.3202 & 0.1619 & \underline{0.6838}   & 0.1624 \\
     \method   & \bf{0.9660} & \bf{0.2277} &  \bf{0.4446}  &  \bf{0.3865}  &  \bf{0.6917}  &  \bf{0.2039}   \\
          \hline
    \end{tabular}%
\end{table}%

\subsection{Performance Comparison}

We first compare models without view attention, as most baselines do not differentiate the importance of views. 

\stitle{Without view attention}
Table~\ref{tab:MICROS_performance} reports the performance comparison between \method\ and baselines without view attention,
which treat all views uniformly.
Our model \method\ achieves the best performance consistently, outperforming the runner-up by up to 35.9\%, 2.0\% and 15.7\% on the three datasets, respectively.
Among the baselines, no consistent winner emerges, with mvn2vec and MNE being generally competitive.

\begin{figure}[t]
  \begin{subfigure}{.3\linewidth}
    \includegraphics[width=0.8\linewidth]{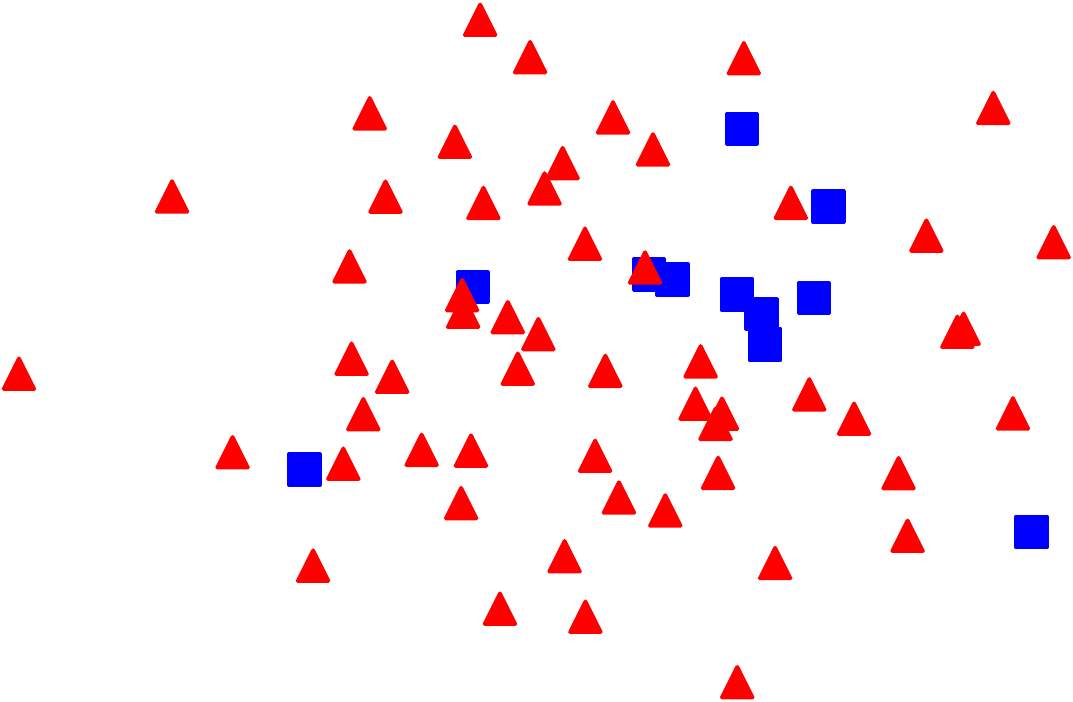}
    \caption{mvn2vec-c}
  \end{subfigure}
  \begin{subfigure}{.3\linewidth}
  \centering
   \includegraphics[width=0.8\linewidth]{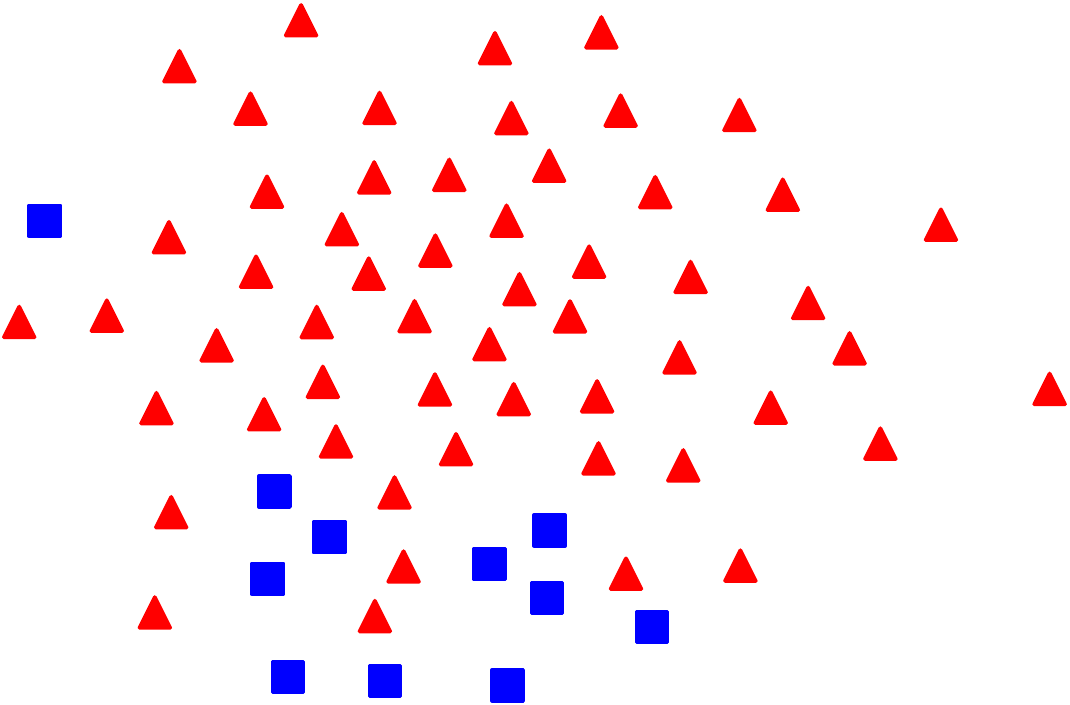}
    \caption{MNE}
  \end{subfigure}
    \begin{subfigure}{.3\linewidth}
   \includegraphics[width=0.8\linewidth]{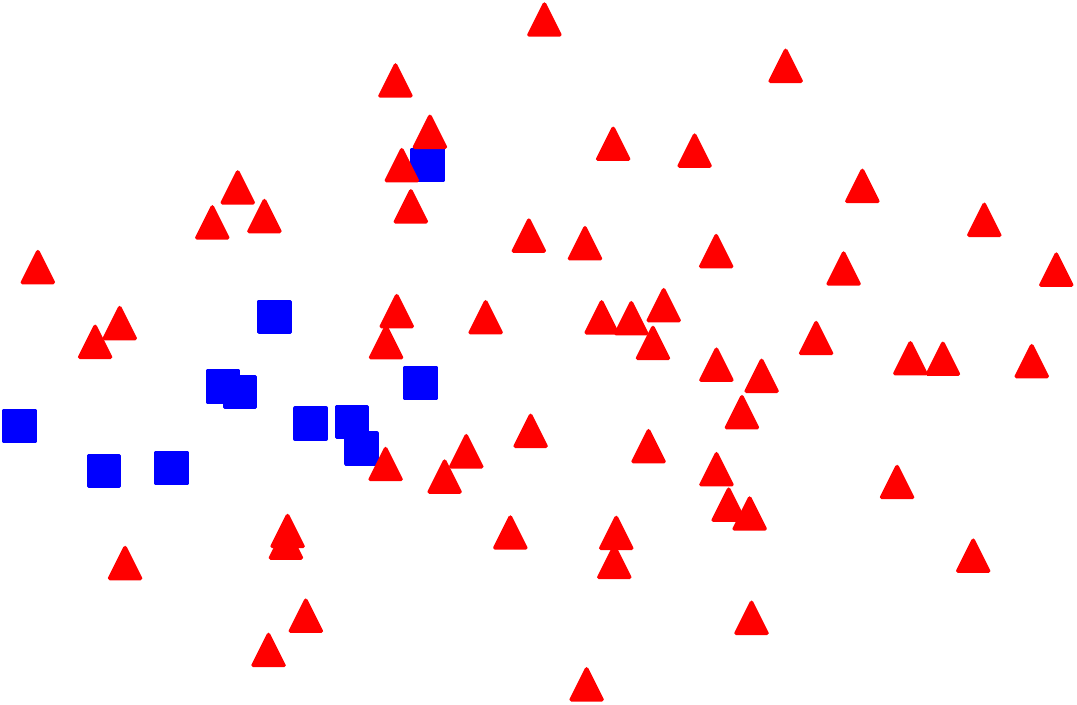}
    \caption{\method}
  \end{subfigure}
      \caption{Visualization of proteins: Alzheimer's positives (\protect\marksymbol{square*}{blue}) and negatives (\protect\marksymbol{triangle*}{red}).
      Best viewed in color.}
            \label{fig:clustering3}
\end{figure}

We further visualize the embeddings generated by \method\ and two competitive baselines in Fig.~\ref{fig:clustering3} on the Alzheimer's dataset. As the dataset is highly skewed with very few positive nodes, for a clearer visualization we uniformly sampled five times as many negative nodes as the positive nodes, while included all positive nodes.
The plot of mvn2vec-c is visibly inferior, with several dispersed positive nodes  far away from the others (which is substantial given that this is a skewed dataset with very few positives). MNE is improved with only one dispersed positive node, although the remaining positives are only  loosely clustered.
In contrast, \method\ forms a relatively dense and well-defined cluster of positives.

\stitle{With view attention} Table~\ref{tab:MICROS-Attention_performance} reports the performance of the attention-based models. 
We make two observations. First, methods with view attention generally outperform their non-attention counterparts shown in Table~\ref{tab:MICROS_performance}.
Second, \methodp\ achieves the best performance in all cases except for ROC-AUC on the highly skewed Alzheimer's dataset. However, on the same dataset \methodp\  outperforms the runner-up in PR-AUC by a substantial margin. 
Note that on highly skewed datasets, the PR curve gives a more informative assessment than the ROC curve \cite{davis2006relationship}.

\begin{table}[tbp]
  \centering
  \caption{Performance evaluation on three datasets with view attention (bold: best; underline: runner-up).}
    \label{tab:MICROS-Attention_performance}%
    \begin{tabular}{l|c|c|c|c|c|c}
    \hline
          & \multicolumn{2}{c|}{Alzheimer's} & \multicolumn{2}{c|}{LinkedIn} & \multicolumn{2}{c}{YouTube} \\
    \hline
    \multicolumn{1}{l|}{ Metric } & \multicolumn{1}{c|}{ ROC-AUC     } & \multicolumn{1}{c|}{ PR-AUC      } & \multicolumn{1}{c|}{ Micro-F    } & \multicolumn{1}{c|}{ Macro-F    } & \multicolumn{1}{c|}{ ROC-AUC      } & \multicolumn{1}{c}{ PR-AUC       } \\
    \hline
     MVE$^+$  & 0.8693 & 0.0067 &  \underline{0.4480}   &  \underline{0.3414}    & 0.6410 & 0.1544 \\
    \hline
     GATNE$^+$  & \bf{0.9719}   & \underline{0.1275}   & 0.3258 & 0.1618 &  \underline{0.6813}   & \underline{0.1597}   \\
    \hline
      \methodp   &  \underline{0.9226} &  \bf{0.3380} & \bf{0.4538}   & \bf{0.3905} &   \bf{0.7214} &  \bf{0.2422} \\
    \hline
    \end{tabular}%
\end{table}%

\subsection{Model Analysis}
\label{sec:expt:param}

We further investigate various aspects of our models, including model ablation, parameter sensitivity, convergence and scalability.

\begin{figure}[t]
	    \centering
	\begin{subfigure}[b]{.25\columnwidth}
	 \centering
	\includegraphics[scale=0.5]{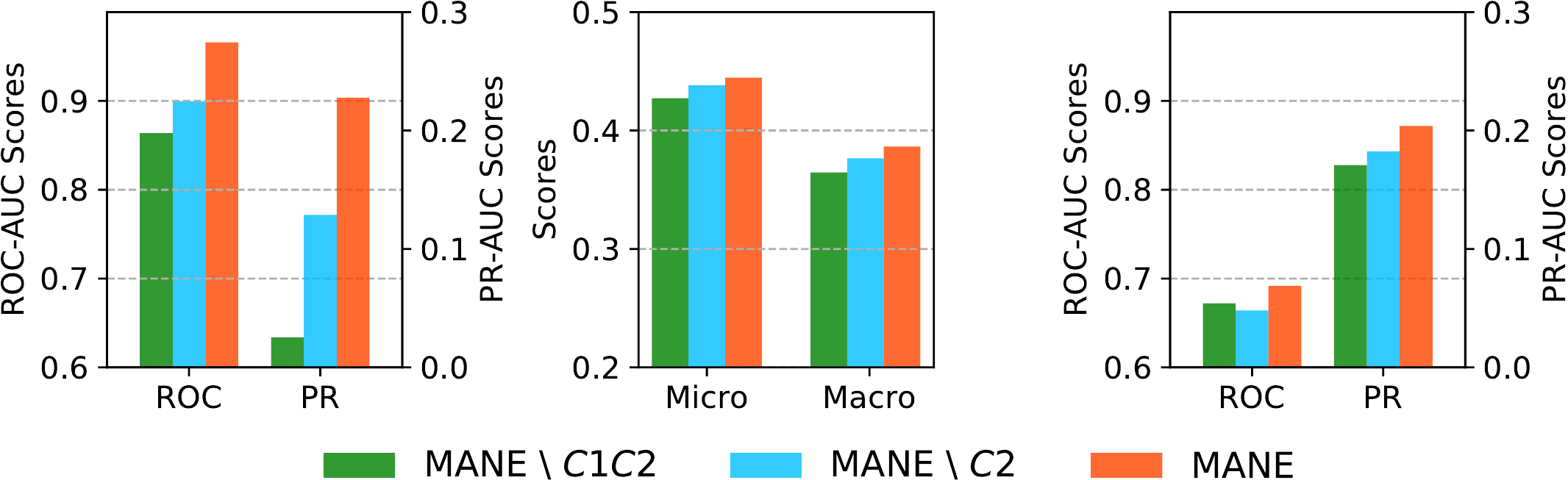}
	\caption{Alzheimer's} 
	\end{subfigure}
	\begin{subfigure}[b]{.26\columnwidth}
	\caption{LinkedIn} 
	\end{subfigure}
	\begin{subfigure}[b]{.27\columnwidth}
	\caption{YouTube} 
    \end{subfigure}
    \caption{Impact of first- and second-order collaboration.}
	\label{fig:ablation}
\end{figure}

\begin{figure}[t]
\centering
    \begin{subfigure}{.3\columnwidth}
    \centering
	\includegraphics[scale=0.65]{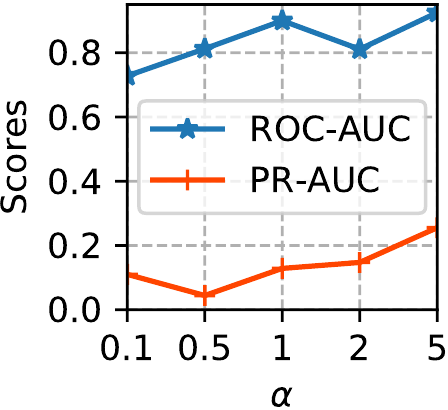}\\[1.5mm]
	\includegraphics[scale=0.65]{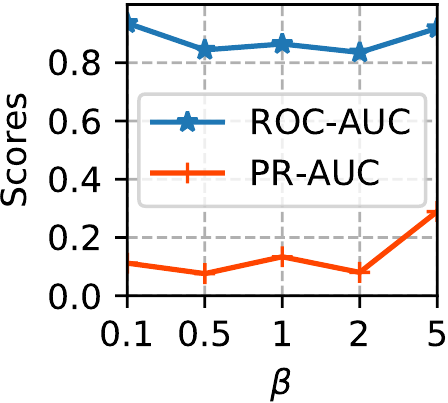}
	\caption{Alzheimer's} 
    \end{subfigure}
	\begin{subfigure}{.3\columnwidth}
	\centering
	\includegraphics[scale=0.65]{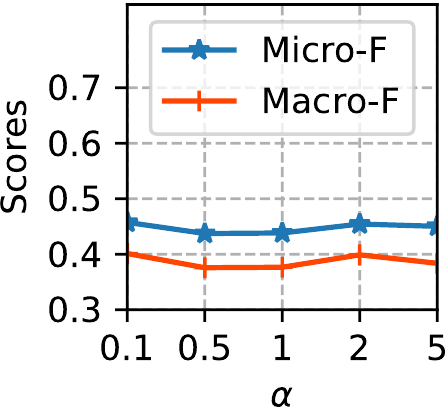}\\[1.5mm]
	\includegraphics[scale=0.65]{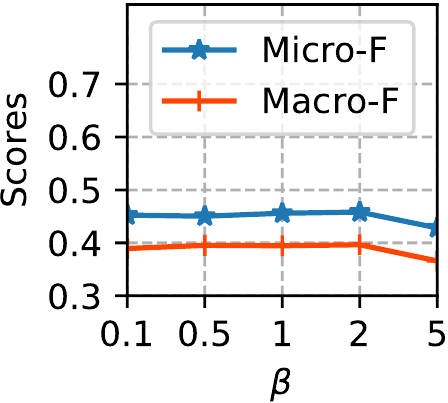}
	\caption{LinkedIn} 
	\end{subfigure}
	\begin{subfigure}{.3\columnwidth}
	\centering
	\includegraphics[scale=0.65]{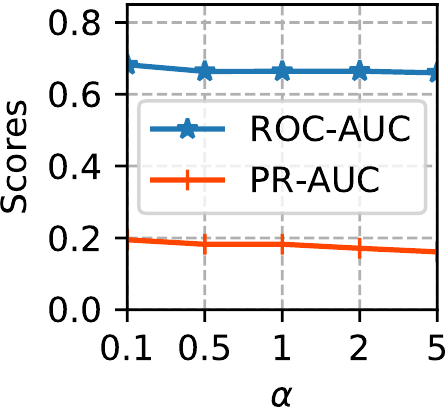}\\[1.5mm]
	\includegraphics[scale=0.65]{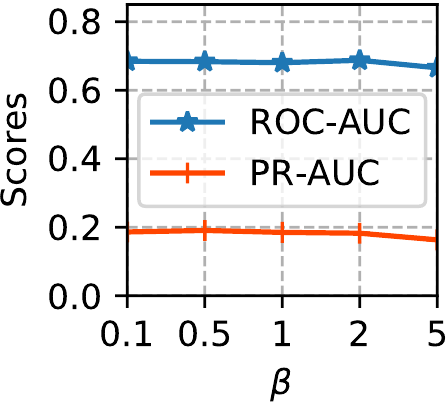}
	\caption{YouTube} 
	\end{subfigure}
    \caption{Impact of hyperparameters $\alpha$ and $\beta$.}
	\label{fig:alpha}
	
\end{figure}

\stitle{Ablation and impact of $\alpha,\beta$}
We examine the contribution from each order of collaboration in Eq.~\eqref{eq:overall_loss}. Specifically, we evaluate (i) \method$\backslash$C1C2 without both orders of collaboration, \ie, $\alpha=\beta=0$; (ii) \method$\backslash$C2 without the second order, \ie, $\alpha=1,\beta=0$.
As Fig.~\ref{fig:ablation} shows, \method\ outperforms \method$\backslash$C2 on all three datasets, validating that the novel second-order collaboration is beneficial. 
In particular, the effect of the second-order collaboration is the largest on Alzheimer's, which is consistent with our analysis in Table~\ref{tab:pairwiseRatio}.
Similarly, \method$\backslash$C2 also consistently outperforms \method$\backslash$C1C2. Overall, the results justify the proposed unification of the three characteristics.
For a finer-grained analysis, we further vary $\alpha$ and $\beta$ in Fig.~\ref{fig:alpha}. The performance is generally robust and stable in the range $[0.5, 2]$ for both parameters.

\begin{figure}[!t]
\centering
    \begin{subfigure}{.3\columnwidth}
    \centering
	\includegraphics[scale=0.65]{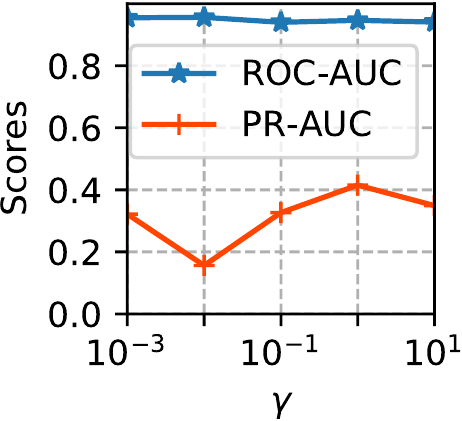}
	\caption{Alzheimer's} 
    \end{subfigure}
	\begin{subfigure}{.3\columnwidth}
	\centering
	\includegraphics[scale=0.65]{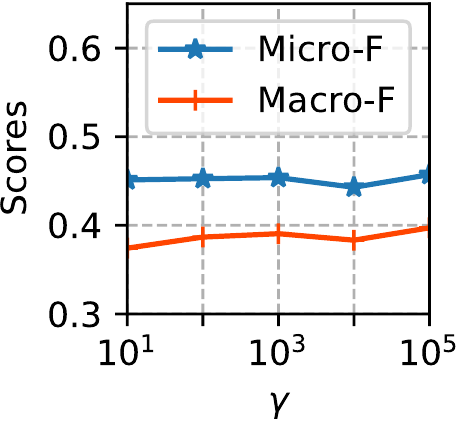}
	\caption{LinkedIn} 
	\end{subfigure}
	\begin{subfigure}{.3\columnwidth}
	\centering
	\includegraphics[scale=0.65]{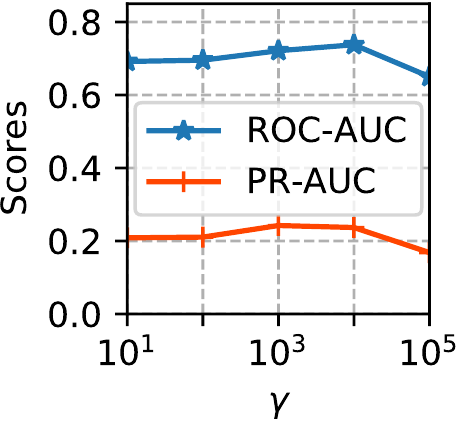}
	\caption{YouTube} 
	\end{subfigure}
    \caption{Impact of hyperparameter $\gamma$.}
	\label{fig:hyp_att}
	
\end{figure}

\stitle{Impact of $\gamma$}
As discussed earlier, on highly skewed Alzheimer's we typically prefer a small $\gamma$, to reduce dependence on the skewed supervision. 
As Fig.~\ref{fig:hyp_att}(a) shows, the range $[0.1,10]$  exhibits robust performance. 
On the other hand, on more balanced datasets YouTube and LinkedIn, a larger $\gamma$ is often useful. As Fig.~\ref{fig:hyp_att}(b) and (c) show, the range $[10^2,10^4]$ gives stable outcomes.

\begin{figure}[!t]
	\centering
	\begin{subfigure}[b]{.45\linewidth}
	\centering
	\includegraphics[scale=0.6]{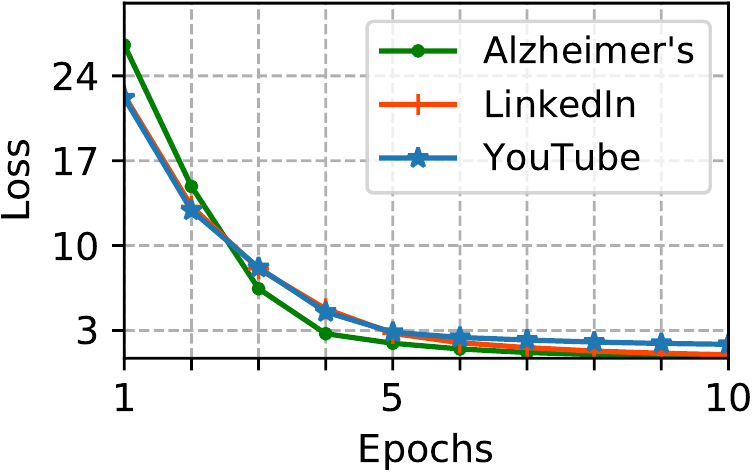}
	\caption{Loss of \method.} 
	\end{subfigure}
	\begin{subfigure}[b]{.45\linewidth}
	\centering
		\includegraphics[scale=0.6]{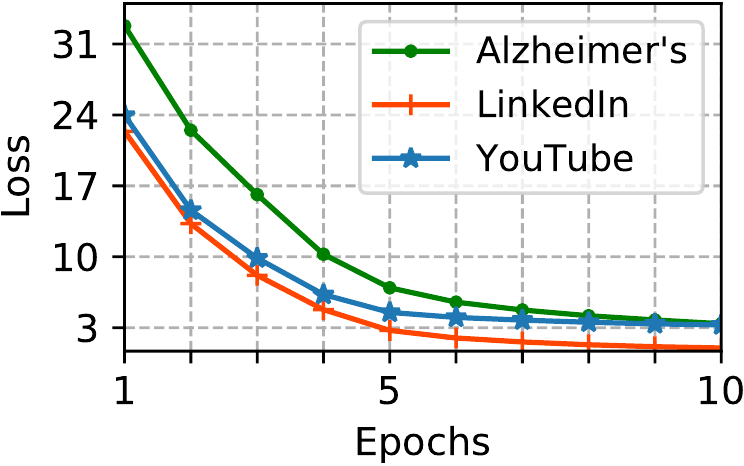}
	\caption{Loss of \methodp.} 
    \end{subfigure}
    \caption{Convergence analysis.}
	\label{fig:convergence}
\end{figure}

\stitle{Convergence analysis}
Fig.~\ref{fig:convergence} shows that our proposed objective functions for both \method\ and \methodp\ converge quickly, typically between 5 and 8 epochs. \methodp\ often has a higher loss as it entails an additional attention-based loss component.  
The actual evaluation scores also demonstrate a similar convergence pattern, which are omitted for brevity.
The fast convergence implies that the different loss components work well together. 

\begin{figure}[!t]
	\centering
	\begin{subfigure}[b]{.49\linewidth}
	\centering
	\includegraphics[scale=0.6]{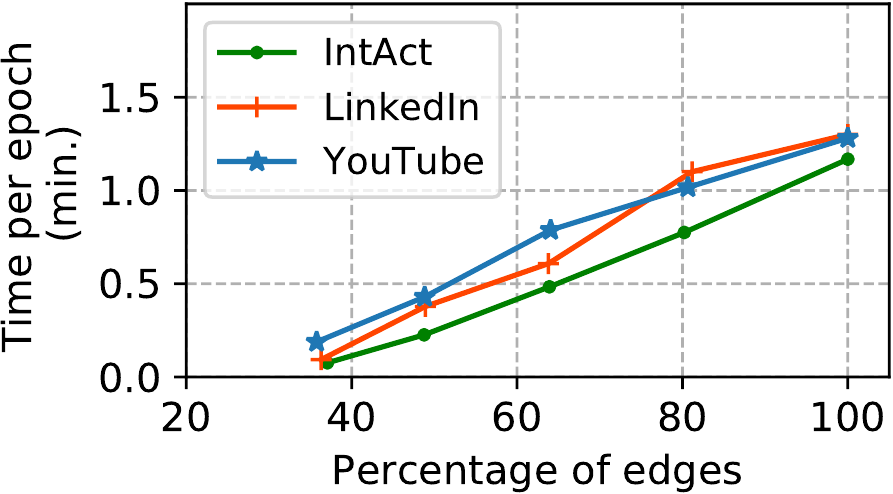}
	\caption{\method.} 
	\end{subfigure}
	\begin{subfigure}[b]{.49\linewidth}
	\centering
		\includegraphics[scale=0.6]{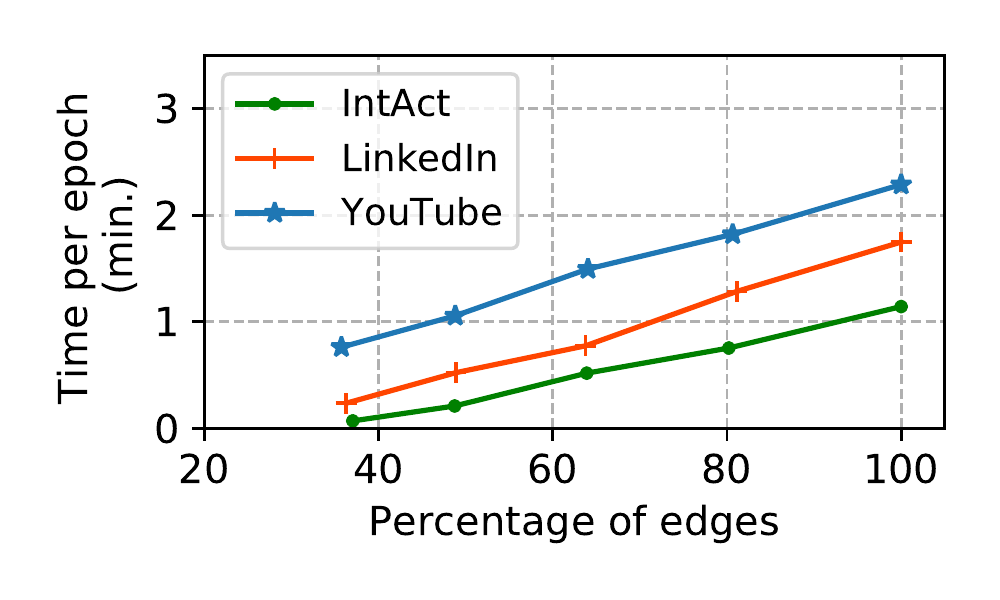}
	\vspace{-3.8mm}
	\caption{\methodp.} 
    \end{subfigure}
    \caption{Scalability analysis.}
	\label{fig:scalability}
\end{figure}

\stitle{Scalability study}
We sampled differently sized multi-view networks from the original datasets. In Figure~\ref{fig:scalability}, we plot the time taken per epoch of training against the total number of edges in all views. 
We observe that both \method\ and \methodp\ scale linearly in the total number of edges, which is consistent with our complexity analysis in Sect.~\ref{sec:model:algorithm}. Moreover, on the IntAct and LinkedIn datasets, we observe that \methodp\ incurs a training duration similar to or slightly higher than \method, as the size of training data on these two datasets are much smaller than the total number of edges.
On the other hand, on the YouTube dataset, \methodp\ incurs a noticeably higher training duration than \method, due to its considerably larger training set. 



\section{Conclusion}
\label{sec:conclusion}
In this paper, we investigated 
an important problem of multi-view network embedding. In addition to the traditional characteristics of diversity and (first-order) collaboration, we leveraged on the newly discovered second-order collaboration. 
We systematically unified the three characteristics with three categories of node pairs, and proposed \method\ for multi-view collaborative network embedding. We further extended it to capture node-wise view importance using the attention mechanism in \methodp. Finally, we evaluated the performance of our proposed approaches, and demonstrated their promising results in comparison to an extensive range of state-of-the-art baselines.
%



\bibliographystyle{ACM-Reference-Format}
\bibliography{ref}



\end{document}